\pdfoutput=1

\documentclass[11pt]{article}

\usepackage[preprint]{acl}

\usepackage{times}
\usepackage{latexsym}

\usepackage[T1]{fontenc}

\usepackage[utf8]{inputenc}

\usepackage{microtype}

\usepackage{inconsolata}
\usepackage{cleveref}
\usepackage{tabulary}

\usepackage{graphicx}
\usepackage{booktabs}
\usepackage{multirow}
\usepackage{pythonhighlight}
\usepackage{comment}
\usepackage{listings}
\usepackage{placeins}

\colorlet{punct}{red!60!black}
\definecolor{background}{HTML}{EEEEEE}
\definecolor{delim}{RGB}{20,105,176}
\colorlet{numb}{magenta!60!black}

\lstdefinelanguage{json}{
    basicstyle=\normalfont\ttfamily,
    numbers=left,
    numberstyle=\scriptsize,
    stepnumber=1,
    numbersep=8pt,
    showstringspaces=false,
    breaklines=true,
    frame=lines,
    backgroundcolor=\color{background},
    literate=
     *{0}{{{\color{numb}0}}}{1}
      {1}{{{\color{numb}1}}}{1}
      {2}{{{\color{numb}2}}}{1}
      {3}{{{\color{numb}3}}}{1}
      {4}{{{\color{numb}4}}}{1}
      {5}{{{\color{numb}5}}}{1}
      {6}{{{\color{numb}6}}}{1}
      {7}{{{\color{numb}7}}}{1}
      {8}{{{\color{numb}8}}}{1}
      {9}{{{\color{numb}9}}}{1}
      {:}{{{\color{punct}{:}}}}{1}
      {,}{{{\color{punct}{,}}}}{1}
      {\{}{{{\color{delim}{\{}}}}{1}
      {\}}{{{\color{delim}{\}}}}}{1}
      {[}{{{\color{delim}{[}}}}{1}
      {]}{{{\color{delim}{]}}}}{1},
}

\usepackage[normalem]{ulem}
\useunder{\uline}{\ul}{}
\newcolumntype{L}{>{\arraybackslash}m{10cm}}

\title{Rule-Guided Feedback: Enhancing Reasoning by Enforcing Rule Adherence in Large Language Models}

\author{A\"issatou Diallo$^1$\thanks{ Corresponding author: \texttt{a.diallo@ucl.ac.uk}\\}, Antonis Bikakis$^2$, Luke Dickens$^2$, Anthony Hunter$^1$, Rob Miller$^2$ \\
$^1$Department of Computer Science \\
$^2$Department of Information Studies     \\
University College London, United Kingdom \\}

\begin{document}
\maketitle
\begin{abstract}
In this paper, we introduce Rule-Guided Feedback (RGF), a framework designed to enhance Large Language Model (LLM) performance through structured rule adherence and strategic information seeking. RGF implements a teacher-student paradigm where rule-following is forced through established guidelines. Our framework employs a Teacher model that rigorously evaluates each student output against task-specific rules, providing constructive guidance rather than direct answers when detecting deviations. This iterative feedback loop serves two crucial purposes: maintaining solutions within defined constraints and encouraging proactive information seeking to resolve uncertainties. We evaluate RGF on diverse tasks including Checkmate-in-One puzzles, Sonnet Writing, Penguins-In-a-Table classification, GSM8k, and StrategyQA. Our findings suggest that structured feedback mechanisms can significantly enhance LLMs' performance across various domains.
\end{abstract}

\section{Introduction}
Large Language Models (LLMs) \cite{brown2020language,openai2303gpt} have gained remarkable performance across various tasks due to their potential reasoning ability.
As the capabilities of large language models (LLMs) expand, they are increasingly utilized in real-world settings characterized by uncertainty and ambiguity. In particular, recent work
aims to develop LLM agents or assistants \cite{xi2023rise,park2023generative} that navigate interactive environments by effectively completing tasks. In this context, rule-following is beneficial as it provides a set of guidelines to enhance rational decision-making and judgments. By following these rules, the agent gets established methods for reasoning and at the same time they have guidance to recognize and avoid reasoning errors. 

\begin{figure}[h]
    \centering
    \includegraphics[width=0.85\linewidth]{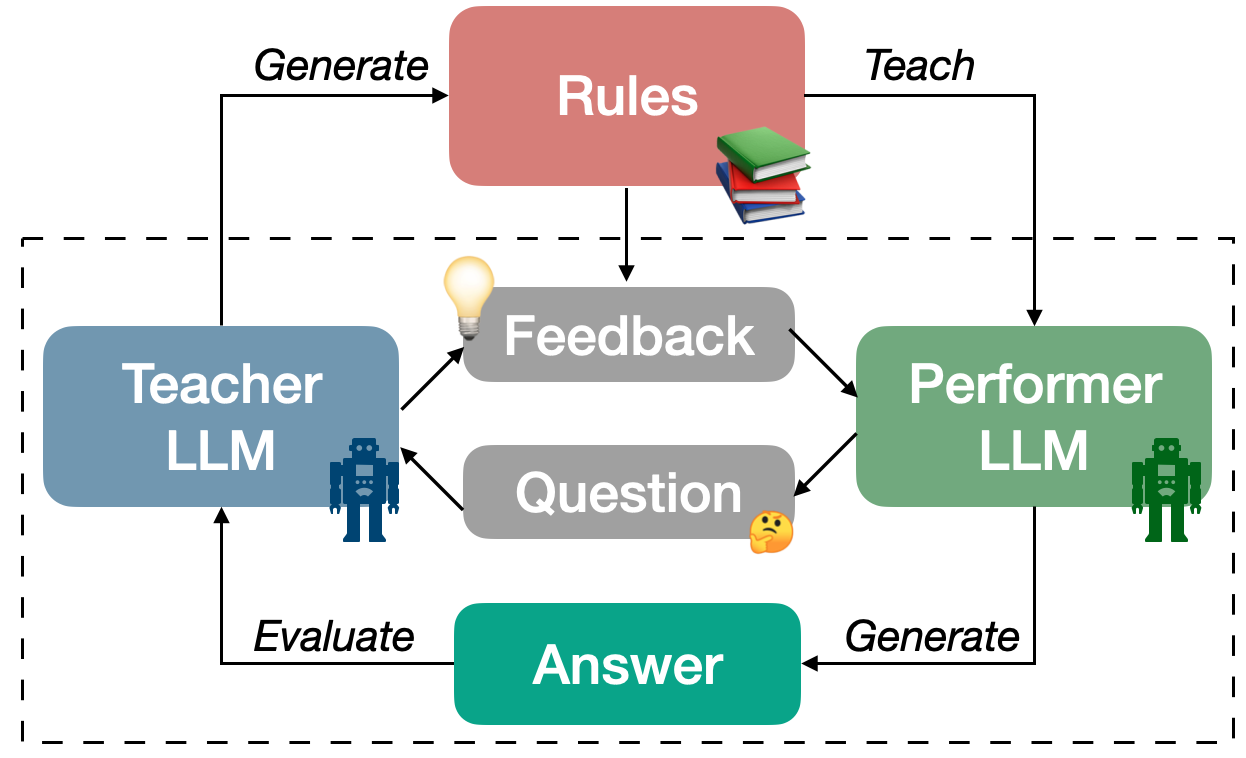}
    \caption{Rule-Guided Feedback framework: Teacher LLM generates rules, provides feedback, and evaluates. Performer LLM generates new answers or questions based on rules and feedback.}
\label{fig:rgf}
\vspace{-0.1cm}
\end{figure}

This is further motivated by the fact that, for many reasoning problems, decomposing and conducting reasoning steps are not sufficient to solve the problem, since one may need domain-specific knowledge to generate correct
steps. Studies  \cite{zhu2023large,wang2024tpd} showed that feedback from rules can significantly boost the performance of baseline prompting methods on different types of reasoning. However, LLMs face difficulties in adhering to rules \cite{mesko2023imperative,lu2024sofa,luo2024asgea}. 

Inspired by these points, we introduce Rule-Guided Feedback, a framework designed to enhance LLM performance by enforcing strict rule adherence and promoting strategic information seeking. Using the dual-agent architecture, our approach employs two distinct roles: the \textit{Performer}, responsible for generating potential solutions based on a provided set of task-related rules, while a \textit{Teacher} rigorously evaluates each output for rule adherence and offers constructive feedback or corrective guidance when deviations are detected. This iterative feedback loop not only ensures that the Performer's outputs remain within the defined constraints but also encourages the model to seek necessary information proactively to resolve uncertainties. The continual dialogue between the Performer and the Teacher additionally helps avoiding the reinforcement of errors from the Performer. The framework is illustrated in \Cref{fig:rgf}.

\textbf{Contributions}: (i) we introduce Rule-Guided Feedback, a plug-and-play approach enabling LLMs to explicitly reason while adhering to the general rules of the task and to seek clarifications to reduce their uncertainty. (ii) Experiments show that our approach improves the success rate of GPT-4 compared with direct prompting achieving top performance on both three challenging tasks.

\section{Approach}
\paragraph{Problem Formulation}

Our framework consists of two agents: a Performer tasked with solving a reasoning problem, and a Teacher who evaluates and guides the Performer's attempts. Each task is defined by a set of natural language rules, as exemplified in \Cref{sec:rules}. The interaction proceeds in turns, where the Performer proposes a solution and the Teacher evaluates it against the predefined rules, providing specific feedback on rule violations. The Performer may request clarification about the feedback, but the Teacher must avoid revealing the solution directly. This exchange continues until either the Performer reaches the correct solution or the maximum number of allowed turns is exhausted. Success is achieved only when the Performer independently arrives at the correct solution while adhering to all rules.

\begin{table}[h]
\centering
\resizebox{\columnwidth}{!}{%
\begin{tabular}{L}
\toprule
\textbf{Checkmate In One} \\ \midrule
\textbf{Ground Truth}: 'Qxe7\#' \\
\textbf{Question:} 1. e4 e5 2. Nf3 d6 3. d4 exd4 4. Nxd4 Nf6 5. Nc3 Qe7 6. Bd3 d5 7. O-O dxe4 8. Re1 Be6 9. Nxe6 fxe6 10. Bxe4 Nxe4 11. Nxe4 Nd7 12. Bg5 Qb4 13. Qg4 Qd4 14. Qxe6+ Be7 15. \\
\midrule
\multicolumn{1}{c}{------ DIALOGUE START ------} \\
\midrule
\textbf{Performer}: The answer is Nf6\#. \\
\midrule
\textbf{Teacher}: Invalid Solution. The proposed move Nf6\# does not result in a checkmate. This is because the knight move to f6 does not put the black king in a position [...]This violates Rule 3. Please review the position [...]. \\ \midrule
\multicolumn{1}{c}{---------- 1 -------------} \\ \midrule
\textbf{Performer}: Apologies for the oversight. To clarify, could you confirm if there are any additional pieces or pawns for either side that might influence the checkmate scenario, aside from those involved in the moves listed? \\ \midrule
\textbf{Teacher}: Based on the moves provided and the sequence leading up to the current position, ...\\ \midrule
\multicolumn{1}{c}{---------- 2 -------------} \\ \midrule
\textbf{Performer}: Thank you for the clarification. Given the current understanding of the board and the pieces involved, I will now propose the correct move that results in an immediate checkmate.

The move is: **Qxd7\#** \\ \midrule
\textbf{Teacher}: Invalid Solution.
The proposed move **Qxd7\#** does not result in a checkmate. The move is legal, but it does not checkmate the black king as the king can escape to other squares [...] This violates Rule 3.
Please review the position [...]. \\ \midrule

\multicolumn{1}{c}{---------- 3 -------------} \\ \midrule
\textbf{Performer}: The answer is Qxe7\#. \\ \midrule
\textbf{Teacher}: Valid Solution. \\
\bottomrule
\end{tabular}%
}
\caption{Example of dialogue for Checkmate In One.}
\label{tab:dialogue}
\end{table}

\paragraph{Overview}

Rule-Guided Feedback starts by using a LLM to generate the rules of the task at hand by a short description. These rules are then used by both the Performer and the Teacher as a guide for their respective roles. For the Performer, they act as guidelines for reasoning and for the Teacher they act as a framework for evaluating and providing feedback. \Cref{fig:rgf} depicts the framework. To evaluate, the Teacher uses (1) the rules, (2) the \textit{history of past interactions} $h_i = \{a_1, f_1, a_2, f_2, ..., a_{i-1}, f_{i-1}\}$ where $a$ is an answer (or question) given by the Performer and $f$ is the feedback by the Teacher, (3) the optional \textit{expert verification}, the outcome of an external algorithm evaluating the Performer output. This last component is necessary for some tasks as the Teacher might lack the domain knowledge to assess the correctness of the Performer's answer as it is. An example of dialogues is shown in \Cref{tab:dialogue}.

\section{Experiments}
In our experiments, we concentrate on GPT-4 \cite{achiam2023gpt} for both Performer and Teacher LLMs. We applied the same parameters for all tasks. We set the temperature value at 0, the maximum number of steps allowed is 5 and the threshold determining the window for the Performer to ask clarifying questions is set to 0.7, that is to say that the Performer can ask question within the first and third turn only. Details about the prompts are presented in \Cref{sec:prompts}.

\paragraph{Baselines} We compare Rule-Guided Feedback (RGF) with the following prompting methods:
(i) \textbf{Standard Prompting (SP)}: the LLM is asked to directly yield a response without any specific guiding input-output exemplars or any additional guiding instructions; (ii) \textbf{Zero-shot CoT prompting (CoT)} \cite{kojima2022large}, inspired by \cite{wei2022chain}, this zero-shot prompting approach simply appends “Let’s think step by
step” to the input query, encouraging the model to generate "thoughts" before addressing the problem or task at hand; (iii) \textbf{Expert Prompting (EP)} \cite{xu2023expertprompting}: functions by first designing an expert identity tailored to the input query and then integrating it into the input to generate a response; (iv) \textbf{Tree of Thoughts (ToT)} \cite{yao2024tree}: enables deliberate decisions by considering multiple reasoning paths and self-evaluating choices to determine the next action; (v) \textbf{Meta Prompting (MP)} \cite{suzgun2024meta}, employs a scaffolding technique where the LLM suggest the ideal prompt, revises it before executing said revised prompt, to help improve outcomes.

\paragraph{Tasks} To evaluate the efficacy of our proposed approach, we consider five tasks with rules that are easily identifiable but requiring a degree of algorithmic reasoning, knowledge, commonsense and literary creativity. (i) \textbf{Penguins In A Table}, directly obtained from the BIG-Bench suite \cite{srivastava2023beyond}: the task is
to answer questions about penguins’ attributes based on a given table and additional natural language information; (ii) \textbf{Checkmate-in-One}, from the same suite, where the goal is to probe the ability of language models to play chess in standard algebraic notation (SAN). The input to the model is a sequence of moves such that a next possible move is a checkmate; (iii) \textbf{Shakespearean Sonnet Writing} from \cite{suzgun2024meta}, a novel task where the goal is to write a sonnet following the strict rhyme scheme "ABAB CDCD EFEF GG" and incorporating three provided words verbatim; \textbf{GSM8k} \cite{cobbe2021training}
 and \textbf{StrategyQA} \cite{geva2021did}, respectively for arithmetic and commonsense reasoning.

\paragraph{Evaluation Metrics}
We employ four key metrics to evaluate performance:
(1) \textbf{Accuracy} (\%): For Penguin-In-A-Table and Checkmate-In-One, we use exact match comparison with ground truth. For Sonnet Writing, we verify functional correctness by checking adherence to all sonnet rules (14 lines, rhyme scheme, meter);
(2) \textbf{Mean Conversation Length in Accurate Cases (MCA)}: Average number of exchanges needed to reach a correct solution, indicating efficiency of successful dialogues;
(3) \textbf{Mean Conversation Length (MCL)}: Average number of total exchanges across all attempts, reflecting overall interaction complexity;
(4) \textbf{Dialogue Density (DD)}: Ratio of cases requiring multiple exchanges to total samples, measuring how often clarification or correction is needed ($DD = N_m/N$, where $N_m$ is the number of multi-turn conversations).
These metrics assess both solution quality and the interaction dynamics of our framework. While accuracy measures correctness, MCA and MCL provide insights into efficiency, and DD quantifies the framework's tendency to engage in iterative refinement when needed.

\subsection{Results}

\begin{figure}[h]
    \centering
    \includegraphics[width=0.9\linewidth]{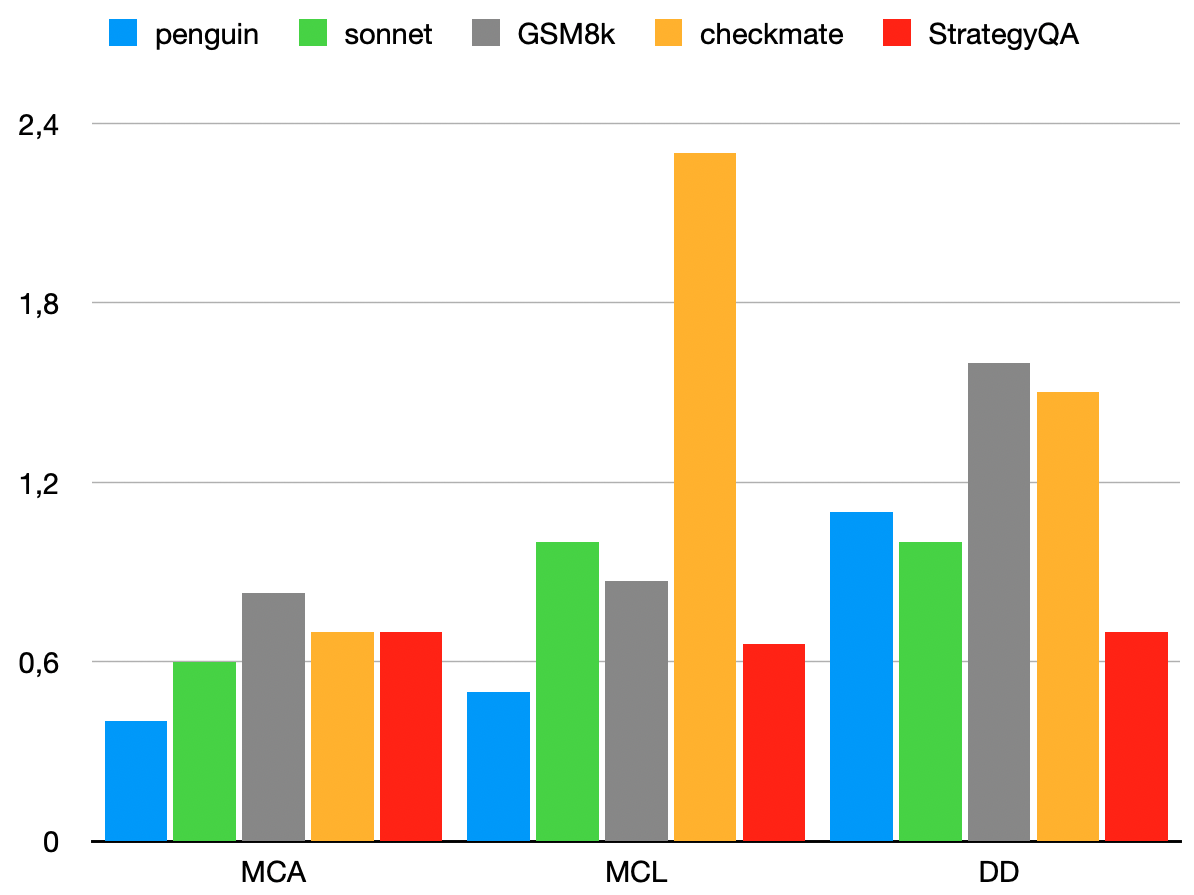}
    \caption{Performance of RGF in terms of Mean Conversation length in Accurate cases (MCA), Mean Conversation Length (MCL) and Dialogue Density (DD). }
\label{fig:rgf_results}
\end{figure}

\Cref{fig:rgf_results} presents the performance of RGF on the five different datasets. 
It is possible to draw a relation between the MCA and the complexity of the problem. In fact, Penguins has an MCA of 0.4 whereas Checkmate had the highest MCA at 2.3, indicating the need for more complex interactions. The same is true for the dialogue density score.  

\begin{table*}[h]
\centering
\resizebox{0.9\textwidth}{!}{%
\begin{tabular}{@{}lcccccc@{}}
\toprule
\multirow{2}{*}{} & \multicolumn{3}{c}{Single Query}                               & \multicolumn{2}{c}{Multi-Query}        & \multirow{2}{*}{\textbf{RGF (Ours)}} \\ \cmidrule(lr){2-6}
                  & \textit{SP} & \textit{GPT-4 + CoT} & \textit{Expert Prompting} & \textit{Meta Prompting} & \textit{ToT} &                                      \\ \midrule
\textit{Checkmate In One}   & 36.4 & 32.8 & 39.6 & {\ul 57.2} & 49.2 & \textbf{62.6} \\
\textit{Penguin In A Table} & 71.1 & 73.6 & 75.8 & {\ul 88.6} & 84.2 & \textbf{97.1} \\
\textit{Sonnet Writing}     & 62.0 & 71.2 & 74.0 & {\ul 79.6} & 68.4 & \textbf{89.4} \\ 
\textit{GSM8k} & 51.0 & 86.2 & 88.0 & 89.5 & {\ul 90.1} & \textbf{93.1} \\ 
\textit{StrategyQA} & 73.0 & 82.1 & 80.0 & 82.5 & {\ul 83.4} & \textbf{94.9} \\ 

\bottomrule
\end{tabular}%
}
\caption{Comparison of Rule-Guided Feedback with previous methods across tasks. We denote the best score in \textbf{bold} and the second-best score is \underline{underlined}. RGF results are average over 3 runs.}
\label{tab:accuracy_table}
\end{table*}

\Cref{tab:accuracy_table} presents a comparative analysis of the baselines prompting methods based on Accuracy. The methods compared include Single Query and Multi-Query. Overall, RGF consistently outperforms the other methods, achieving the highest scores in all three tasks. This suggests that providing language models with explicit rules and guidelines significantly improves their performance on these types of tasks. Generally, Multi-Query methods achieve higher scores. This indicates that setting a dialogue for finding the solution is beneficial.

In conclusion, the results demonstrate the effectiveness of RGF in improving the performance across different tasks. 

\section{Ablation Studies}

We conduct ablation studies on three key components of our Rule-Guided Feedback framework to understand their contributions. \Cref{fig:clar_ques} shows the impact of main components of RGF.
\paragraph{Impact of Clarifying Questions}
Removing the Performer's ability to ask clarifying questions leads to a 15\% decrease in accuracy across tasks on average. This decline is most notable in Checkmate-in-One, where understanding piece positions and movement rules is crucial. The ability to seek clarification helps resolve ambiguities early, preventing cascading errors. 
\paragraph{Impact of Expert Validation}
Without access to explicit rule violation lists, the Teacher's evaluation accuracy drops by 8.5\%, with a 12\% increase in undetected rule violations (falsely claiming the Performer is right or wrong). The expert validation step is crucial for catching edge cases, ensuring consistent feedback quality, and providing additional verification for complex tasks. \Cref{fig:rule_viol} shows the impact of expert verification.
\paragraph{Impact of Feedback Iteration Limit}
We analyze performance across different iteration limits (1, 3, 5, and unlimited). Most successful solutions are reached within 3 iterations, with 5 iterations providing optimal performance (13\% improvement over single iteration). Additional iterations beyond 5 yield minimal benefits (<1\%) while significantly increasing computational cost. The optimal iteration limit varies by task - Checkmate-in-One shows 22\% improvement between 1-3 iterations, while Penguins-in-a-Table typically succeeds within 2-3 iterations. \Cref{fig:num_iter} shows the impact of the number of iterations on accuracy.

\section{Discussion}
While the RGF framework enhances (LLMs) adherence to predefined rules and improves information-seeking efficiency, it is not without limitations. One challenge occurs from the Performer's tendency to persist with previous answers despite receiving corrective feedback from the Teacher. This issue sometimes happens even if both the Performer and the Teacher have the memory of prior interactions, leading to repeated errors and diminished overall effectiveness of the feedback loop. 

Another limitation comes from the Teacher's ability to provide accurate and constructive feedback. The efficacy of the framework relies on the Teacher's capacity to thoroughly understand the problem, accurately identify the Performer's mistakes, and articulate these errors clearly as violated rules. If the Teacher fails to comprehend the issues of the Performer's responses or misinterprets rule violations, the feedback becomes less effective, hindering the Performer's "reasoning".

To address these challenges, we have incorporated an expert verification step following each Performer's response. This addition serves a dual purpose: it verifies the correctness of the Performer's answer and ensures that any feedback provided highlights specific rule violations in a clear and actionable manner. By requiring the Teacher to rephrase feedback in terms of explicit rule violations, we enhance the clarity and utility of the guidance offered to the Performer. This structured approach not only mitigates the Performer's inclination to disregard feedback but also improves the Teacher's role in providing meaningful corrections.

Future work could use a reward system to better align the Performer with the Teacher.

\section{Conclusion}

This paper introduces Rule-Guided Feedback, a framework that improves LLMs' adherence to rules and enhances their information-seeking capabilities. On three datasets RGF increases accuracy rate by 26.5\% on average over direct prompting and 7.9 \% over the best baseline which demonstrates substantial improvements in task completion rates. Future work will focus on refining the feedback integration process and enhancing the Teacher's analytical capabilities to further strengthen the framework's robustness. Ultimately, RGF represents a step forward in providing reliable and intelligent LLM interactions within rule-based environments.

\section*{Limitations}
As stated previously, one limitation of this work consists in the fact that the Performer's occasionally fails to effectively integrate the Teacher's feedback, leading to repeated adherence to incorrect solutions despite available conversational history. Additionally, the framework's reliance on the Teacher's ability to accurately comprehend and identify the Performer's mistakes depends on the domain-knowledge and reasoning ability of the Teacher model. While the inclusion of expert verification helps mitigate these issues by ensuring the correctness of evaluations, there remains a need for enhancing the Performer's responsiveness to feedback and further improving the Teacher's analytical capabilities to consistently deliver precise and actionable guidance. These points are left for future work.

\section*{Ethical Statement}
Throughout the paper, we utilize anthropomorphic language, such as "reason" and "understand" to enhance the clarity of our discourse on LLMs. This is not intended to imply that LLMs possess human-like cognitive abilities.

The generative models are based on pre-trained language models, which may
generate offensive content if prompted with inappropriate inputs. Additionally, concerns may arise regarding potential bias amplification and misuse as per any prompting method for LLMs. This necessitates careful bias detection and mitigation. Additionally, the rules could exploited to "trick" the model, generating harmful or misleading outputs and feedback. Safeguards like fact-checking and verification become crucial, especially in sensitive domains. Responsible deployment and use are fundamental to ensure LLMs benefits are harnessed ethically and its risks are minimized.

\section*{Acknowledgements}

This research was supported by the Leverhulme Trust grant for the project 'Repurposing of Resources: from Everyday Problem Solving through to Crisis Management' (RPG-2021-182). We also thank all reviewers for their insightful feedback.

\bibliography{main}

\appendix

\section{Additional Results}

The following graphs show the quantitative results of the ablation studies.
\begin{figure}[h!]
    \centering
    \includegraphics[width=0.9\linewidth]{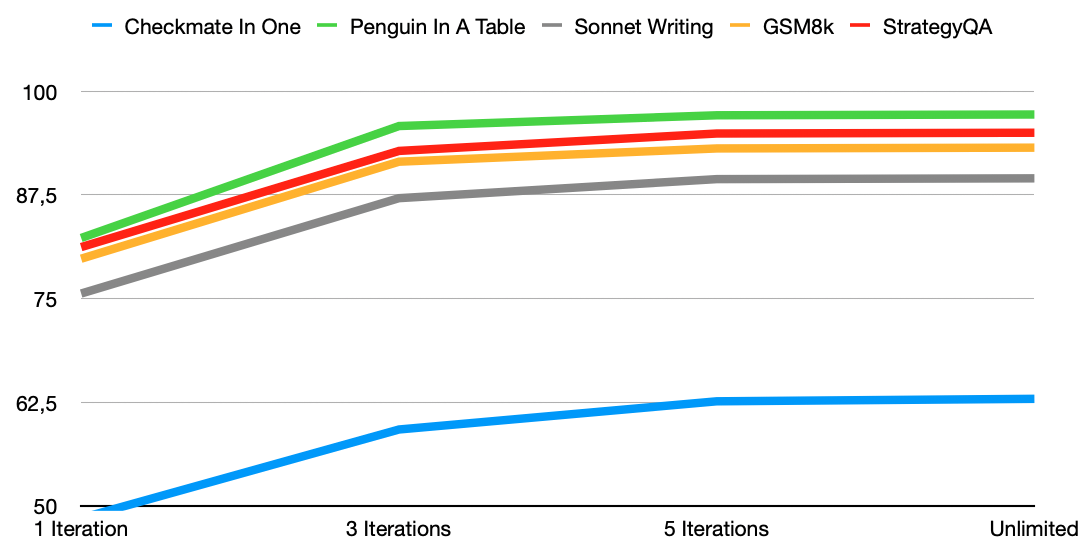}
    \caption{Performance across different iteration limits. Most tasks achieve optimal results within 5 iterations, with diminishing returns thereafter}
\label{fig:num_iter}
\end{figure}

\begin{figure}[h!]
    \centering
    \includegraphics[width=0.9\linewidth]{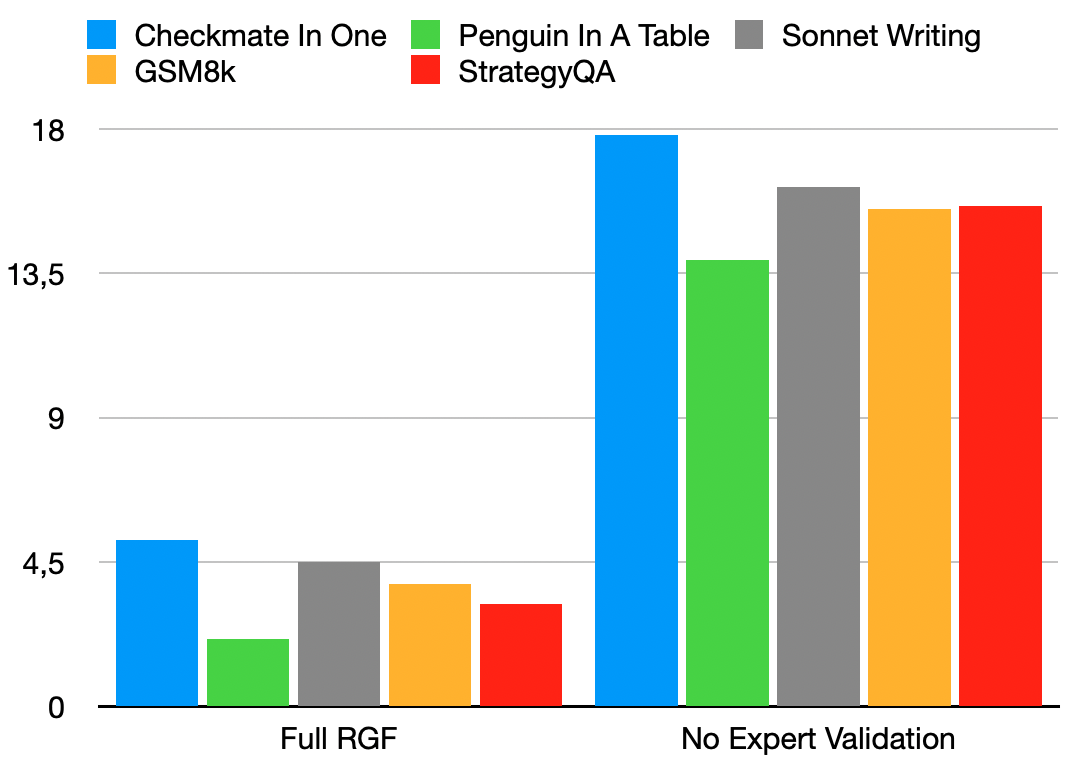}
    \caption{Rule violation rates with and without expert validation. Expert validation reduces rule violations by an average of 12\% across tasks.}
\label{fig:rule_viol}
\end{figure}

\begin{figure}[h!]
    \centering
    \includegraphics[width=0.9\linewidth]{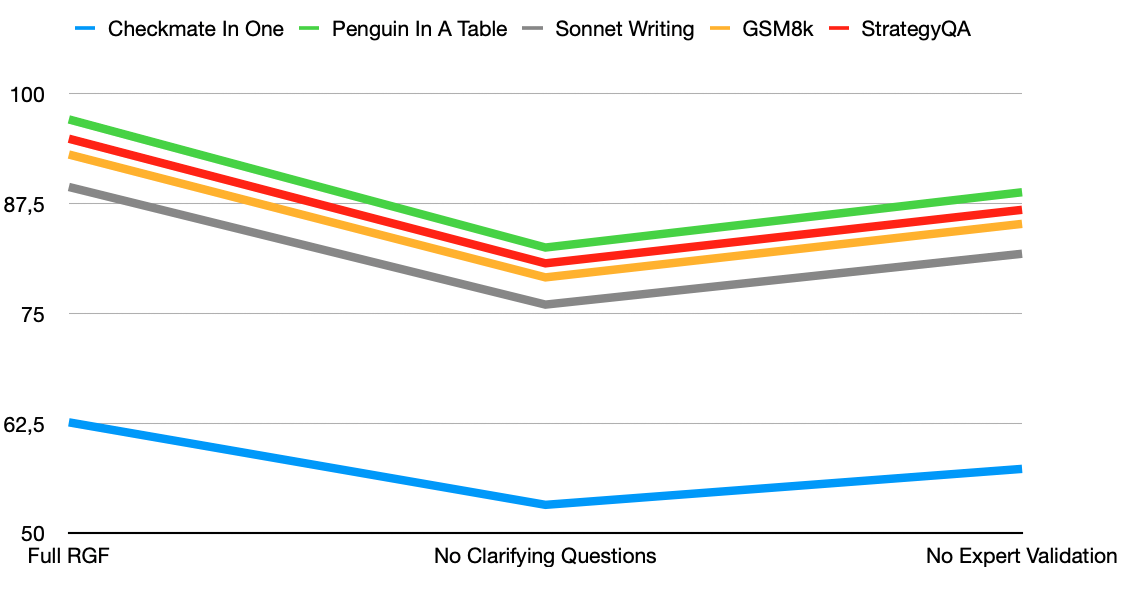}
    \caption{Comprehensive analysis of RGF components. Performance comparison showing relative contribution of each framework component and the impact of removing key components from RGF framework. Removing clarifying questions and expert validation consistently reduces performance across all tasks.}
\label{fig:clar_ques}
\end{figure}

\begin{table*}[h]
\centering
\begin{tabular}{lccc}
\toprule
 & \textbf{Checkmate} & \textbf{Penguin} & \textbf{Sonnet} \\
\midrule
w/o Examiner & 36.4 & 71.1 & 62.0 \\
w/o clarifying questions & 41.4 & 81.7 & 97.1 \\
RGF Performer & 80.6 & 41.5 & 122.2 \\ 
RGF Examiner & 386.1 &	200.2 &	84.5 \\ \midrule
CoT & 556.4	& 25.8 &	176.2 \\
Meta-Prompting & 523.1 & 14.3 & 173.7 \\ 
\bottomrule
\end{tabular}
\caption{Average output token count for RGF (worst case: when max turn is reached)}
\label{tab:model_performance}
\end{table*}

\FloatBarrier

\section{Expert Verification}

Expert verification step is used as tool due to the inherent limitation of the LLMs with some challenging tasks. It is a call to a function to programmatically evaluate the performer answer. It is used in SonnetWriting specifically as GPT-4 couldn’t reliably evaluate these points:

\begin{itemize}
    \item Check if the sonnet adheres to a given rhyming scheme
    \item Check if each line has 10-11 syllables, more precisely, there's some pronunciation of each line with 10-11 syllables.
\end{itemize}

The python libraries used for this evaluation are:
\begin{itemize}
    \item \texttt{pronounce} - a well-known pronunciation dict based on from CMU's pronouncing dictionary
    \item \texttt{syllables} - a Python library for syllable counting
    \item \texttt{pyphen} - a Python wrapper for the hyphenation library
\end{itemize}

Once the Performer outputs an answer, this is passed to the expert verification function and evaluated. Then. this evaluation and the Performer's answer are passed to the Teacher.

\section{Related Work}

\paragraph{Teacher-student framework} The teacher-student framework has traditionally focused on knowledge transfer from larger teacher models to smaller student models \cite{gou2021knowledge}. Early approaches relied on supervised fine-tuning methods \cite{magister2022teaching,shridhar2022distilling}, as demonstrated by Rajani et al. (2019) who generated explanations using GPT \cite{radford2018improving} to enhance BERT models' commonsense reasoning. Similarly, \citet{hendrycks2021measuring} improved mathematical reasoning through fine-tuning on step-by-step solutions, while \citet{ho2022large} leveraged GPT-3 \cite{brown2020language} to generate Chain-of-Thought reasoning steps for training smaller models. However, these approaches face significant limitations: the fine-tuning process is resource-intensive, collecting task-specific data is challenging, and the resulting models often lack cross-task generalization capabilities.
Recent work has shifted towards prompting-based teacher-student frameworks to address these limitations \cite{saha2023can,yu2023characterizing}. Most relevant to our work is Teaching via Principle Discovery (TPD) \cite{wang2024tpd}, which uses a teacher LLM to generate problem-solving instructions and corrective principles based on student errors. While both TPD and RGF leverage teacher-student interactions, they differ fundamentally in their approach and objectives. TPD focuses on discovering principles from errors to improve instruction and training, with the teacher disengaging after principle generation. In contrast, RGF maintains active teacher involvement during inference, using predefined rules to guide real-time reasoning and information seeking. Furthermore, while TPD aims to transfer knowledge between models through principle-based learning, RGF emphasizes structured reasoning through continuous rule adherence and feedback. These distinctions highlight RGF's novel contribution to the field, particularly in scenarios requiring strict rule compliance and dynamic information gathering.

\section{Datasets}
\begin{itemize}
    \item Penguins In A Table: (size: 149) \url{https://github.com/google/BIG-bench/tree/main/bigbench/benchmark_tasks/penguins_in_a_table}
    \item Checkmate In One: (size: 250)  \url{https://github.com/google/BIG-bench/tree/main/bigbench/benchmark_tasks/checkmate_in_one}
    \item Shakesperean Sonnet Writing: (size: 250) \url{https://github.com/suzgunmirac/meta-prompting/blob/main/data/Sonnets-Standard.jsonl}
    \item GSM8k: (size: 1319) \url{https://github.com/openai/grade-school-math}
    \item StrategyQA: (size: 2290) \url{https://github.com/google/BIGbench/tree/main/bigbench/benchmark tasks/strategyqa}
\end{itemize}

\subsection{Task Descriptions}
\textbf{Checkmate In One}: Given a series of chess moves written in Standard Algebraic Notation (SAN), determine the next move that will result in a checkmate.

\textbf{Penguins In A Table}: Given a series of table, interpret the eventual extra modifier, and answer the question correctly.

\textbf{Sonnets}: Write a sonnet that adheres strictly to the specified rhyme scheme ABAB CDCD EFEF GG and includes the given words.

\textbf{GSM8k}: a dataset of 8.5K diverse, multi-step grade school math word problems, solvable by basic arithmetic

\textbf{StrategyQA}: question-answering benchmark focusing on open-domain questions where the required reasoning steps are implicit in the question and should be inferred using a strategy.

\section{Rules}
\label{sec:rules}
\subsection{Checkmate In One}
1. All provided chess moves must be written in Standard Algebraic Notation (SAN).

2. The sequence of moves must accurately reflect a valid game progression up to the current position.

3. The next move must result in an immediate checkmate of the opponent's king.

4. The proposed move must comply with the official rules of chess, including correct piece movement and capturing mechanisms.

5. Only one move should be provided as the solution, representing the checkmating move.

6. The move must be legal in the given position and must not leave the player's own king in check.

7. If the checkmate involves special moves (such as castling, en passant, or pawn promotion), the notation must correctly represent these moves.

8. Disambiguation in SAN must be used correctly when multiple identical pieces can move to the same square (e.g., Nbd7).

9. The solution must demonstrate that no additional moves are required after the proposed move to achieve checkmate.

10. The performer is allowed to ask one clarifying question, but it cannot be a repetition of the original question.

\subsection{Penguins In A Table}

1. Each table must have a header row with clearly labeled columns.

2. Each subsequent row represents one penguin with corresponding attribute values.

3. Attribute names in questions must exactly match those in the table headers.

4. Questions must be answerable solely using the information provided in the table(s).

5. Numerical values (age, height, weight) must be interpreted and compared accurately.

6. Units of measurement (e.g., cm, kg) must be correctly used in both questions and answers.

7. When comparing attributes, ensure that the relationships (e.g., taller than) are correctly determined based on the data.

8. If multiple tables are present, clearly reference the correct table when answering questions about specific penguins.

9. All answers must be clear, concise, and directly derived from the table data without introducing external information.

10. The performer is allowed to ask one clarifying question, but it cannot be a repetition of the original question.

\subsection{Sonnet Writing}
1. The sonnet must consist of exactly 14 lines.

2. Strictly follow the specified rhyme ABAB CDCD EFEF GG scheme throughout all 14 lines.

3. Include all the given words within the sonnet.

4. Adhere to the traditional structure associated with the chosen rhyme scheme (Shakespearean).

5. Ensure that each line contributes coherently to the overall theme or message of the sonnet.

6. Use the given words naturally, without forcing them into the lines.

7. Ensure proper punctuation and capitalization are used to enhance readability and flow.

8. The number of allowed syllables per line is 10 or 11.

9. The performer is allowed to ask one clarifying question, but it cannot be a repetition of the original question.

\subsection{GSM8k}

1. **Identify the Core Mathematical Operations:** Accurately determine the sequence of arithmetic operations (addition, subtraction, multiplication, division, percentages) required to solve the problem.
2. **Break Down the Problem into Steps:** Decompose the problem into smaller, more manageable sub-problems. 
3. **Extract Relevant Information:** Identify and isolate the crucial numerical values and units from the problem statement.
4. **Perform Calculations Accurately:** Execute the identified arithmetic operations correctly, paying close attention to order of operations (e.g., parentheses, exponents, multiplication/division, addition/subtraction).
5. **Handle Units Correctly:** Ensure proper unit conversions and consistency throughout the calculations.
6. **Address Multi-Step Problems:** For problems involving multiple steps, maintain intermediate results and clearly label each step in the solution.
7. **Check for Reasonable Answers:** After obtaining the final answer, perform a quick sanity check to ensure it aligns with expectations and common sense.
8. **Consider Special Cases:** Be mindful of potential edge cases or unusual scenarios described in the problem (e.g., discounts, taxes, varying speeds).
9. **Document the Solution Clearly:** Present the solution in a structured and organized manner, showing all intermediate steps and calculations.
10. **Utilize Visual Aids (Optional):** If helpful, create simple diagrams or tables to visualize the problem and aid in understanding. 

\subsection{StrategyQA}

1.Identify the Main Question: The response must accurately identify the core sub-question within the given StrategyQA question.
2. Recognize Implicit Strategies: The response must explicitly state the underlying reasoning steps required to answer the main question, even if these steps are not explicitly stated in the original question.
3. Leverage Background Knowledge: The response must demonstrate the use of general knowledge and world facts to understand the context and relationships between concepts within the question.
4. Focus on Relevant Evidence: The response must clearly indicate which evidence paragraphs are most relevant to the sub-question and the identified reasoning steps.
5. Reasoning Chain Construction: The response must present a clear and logical chain of reasoning that connects the sub-question, the identified evidence, and the final answer.
6. Answer Justification: The response must include a justification for the answer using the collected "facts" (justification) provided by the question writers.
7. Identify Ambiguous Questions: If the question exhibits ambiguity, the response must acknowledge this ambiguity and explain how the chosen interpretation is supported by the evidence.
8. Counterfactual Reasoning: The response may include a brief discussion of alternative scenarios and how they would impact the answer based on the provided evidence.
9. Identify Out-of-Scope Information: The response must demonstrate an understanding of which information in the evidence paragraphs is irrelevant to the main question or the reasoning steps.
10. Confidence Calibration: The response must include a brief statement indicating the confidence level in the provided answer and justification, based on the strength and clarity of the supporting evidence.

\section{Prompts}
\label{sec:prompts}
\subsection{Checkmate In One}
\onecolumn
\begin{python}
#Performer Prompts
performer_prologue = '''Let us play chess. You are the an expert chess player who will analyze the provided chess moves and determine the next move that will result in a checkmate.

You will suggest a single chess move that, when played, will checkmate the opponent. The move must result in an immediate checkmate.
Take into account the Teacher's feedback and do not repeat wrong moves.

Previous moves:
{input}

First, write the position of all pieces on the board based on the previous moves.

The rules you must follow are:
{rules}

Let us begin. Please propose your next move or ask a clarifying question.
'''

urge_prompt = """Note that you should guess what the answer is exactly from now on without repeating your mistakes.
The answer must start with 'The answer is ..."""  

# Teacher Prompts
examiner_prologue = '''Let us review the proposed move to ensure it results in a checkmate.
Here are the previous moves:
{input}

You are the Examiner responsible for verifying the Performer's suggested move and compare it to the correct move.

Here is the correct move: {target}. Do not mention it in your dialog with the Performer.

Please check the following rules:
{rules}

If the move is correct and results in a checkmate, respond with: "Valid Solution" otherwise "Invalid Solution".

If the move does not result in a checkmate or violates any rules, specify why it is incorrect, which rule it violates and provide constructive feedback for guidance but do not suggest the correct move.

Let us begin. Here is the proposed move by the Performer:
'''
\end{python}

\subsection{Penguins In A Table}

\begin{python}
# performer (guesser)
performer_prologue = '''Let us engage in an Animal QA session. The goal is to answer questions about a table of penguins and their attributes. You are the Performer who will propose potential answers based on the provided tables containing information about various animals.
Here is the table information:
{input}

You will suggest one possible answer at a time based on the question. Each answer should be directly supported by the data in the tables.

You must follow these rules:
{rules}

Let us begin. Please provide your first answer to the question. Write the solution in this following format "The answer is ... "  
'''

urge_prompt = """Note that you should guess what the answer exactly is from now on.
The prompt must start with 'The answer is ...'"""

# teacher (examiner)
examiner_prologue = '''Let us engage in an Animal QA session. You are the Teacher responsible for verifying the Performer's proposed answers.
{input}
Please verify the following:
{rules}

Compare the Performer answer with the ground truth answer {target}.
If the Performer's answer is correct, respond with: "Valid Solution" otherwise respond with "Invalid Solution".

If there are any discrepancies, specify which rules are incorrect and provide constructive feedback for correction but the correct answer must not be given in your answer. If the performer asks a clarifying question, reply to the question but do not give the correct answer and rewrite the original question.
Let us begin. 
'''
    
\end{python}

\subsection{Sonnet Writing}

\begin{python}
# Performer Prompts
performer_prologue = '''You're an expert poet. Let us compose a sonnet adhering strictly to the specified rhyme scheme and containing each of the following words verbatim {input}.

Here are the rules you must follow:
{rules}

You will write a 14-line sonnet that follows the rhyme scheme ABAB CDCD EFEF GG. Ensure that all the provided words are seamlessly integrated. You can choose the theme and the emotion.

Let us begin. Please compose your first sonnet incorporating the provided words.
'''

urge_prompt = """Note that you should write what the solution is from now on.
The prompt must start with 'The answer is ..."""  # Optionally, can be customized or kept empty based on specific needs

# Teacher Prompts
examiner_prologue = '''Let us review the sonnet that should contain each of the following words verbatim {input}, to ensure it adheres to the specified guidelines here:
{rules}

You are the Examiner responsible for verifying the Performer's proposed sonnet.
An expert may help you verify the performer response and will give a statement starting with "Expert verification: ..."

If the sonnet is correct, respond with: "Valid Solution", otherwise "Invalid Solution".

If there are any discrepancies or errors, specify which rules are incorrect and provide constructive feedback for correction but do not rewrite the sonnet. Point at the lines containing the mistakes.

Let us begin. Here is the proposed sonnet:
\end{python}

\subsection{GSM8k}

\begin{python}
# Performer (Solver) Prompts
performer_prologue = '''Welcome! You are a problem solver tasked with finding solutions to a specific scenario described below.

Scenario:
{input}

Analyze the problem and develop a step-by-step solution that achieves the stated objective. Be sure to adhere to the following rules:

{rules}

Let us begin. Please outline your first solution or ask a clarifying question. Write you answer as:
'The answer is ...'
'''

urge_prompt = """Note that you should guess what the answer exactly is from now on.
The prompt must start with 'The answer is ...'"""

# Examiner (Evaluator) Prompts
examiner_prologue = '''Let us evaluate the proposed solution to ensure it effectively solves the problem stated as {target}.
Below are the details of the problem provided:
{input}

You are the Examiner tasked with verifying the Performer's suggested solution against the correct solution.

Here is the correct solution: {target}. Do not reveal it during your conversation with the Performer.

Consider the following rules:
{rules}

If the Performer's solution is correct, respond with: "Valid Solution". If not, respond with "Invalid Solution".

If the proposed solution is inadequate or breaks any rules, explain why it is incorrect, which rule it violates, and provide guiding feedback but refrain from suggesting the correct solution.

Let us begin. Here is the problem solver's proposed solution:

'''
\end{python}

\subsection{StrategyQA}

\begin{python}

# Performer (Strategist) Prompts
performer_prologue = '''Welcome! You are a strategist tasked with devising a plan to achieve specific targets given a particular scenario.

First, review the current question:
{input}

Be mindful of the rules outlined and ensure compliance when giving your answer:

{rules}

Let us begin. Please provide your first answer to the question. Write the solution in this following format "The answer is ... [yes/no]
'''

urge_prompt = """Remember that your response should precisely identify the proposed strategy from now on without repeating past mistakes.
The answer must start with 'The proposed strategy is ...'"""

# Examiner (Evaluator) Prompts
examiner_prologue = '''Let us review the proposed strategy to ensure it meets the objectives for achieving {target}.
Here are the details of the scenario provided:
{input}

You are the Examiner responsible for verifying the Performer's proposed strategy and comparing it to an optimal one.

Here is the expected strategy: {target}. Please do not disclose it during your interaction with the Performer.

Please consider the following rules:
{rules}

If the answer is correct, respond with: "Valid Solution". If not, state "Invalid Solution".

Should the proposed answer fall short or violate any rules, clarify why it is incorrect, which rule it breaches, and provide constructive feedback for guidance but refrain from suggesting the correct answer.

Let us begin. Here is the Performer's proposed answer:

'''
\end{python}

\section{Baselines}

\paragraph{CoT} \texttt{Let's think step by step}
\paragraph{EP} 
\begin{itemize}
    \item \textbf{meta prompt}: For each instruction, write a high-quality description about the most capable and suitable agent to answer the instruction. In second person perspective.
    \item \textbf{Checkmate In One}: You are a seasoned chess grandmaster. ... Your task is to provide a single, definitive move that will deliver the final blow.
    \item \textbf{Sonnet}: You are a master of the sonnet form, a poet ...Your task is to compose a sonnet that is both technically sound and emotionally resonant.
    \item \textbf{Penguins}: You are a knowledgeable zoologist ... Your goal is to demonstrate your expertise in animal science and provide insightful answers that showcase your understanding of the penguin species.
\end{itemize}

\paragraph{Meta Prompting}
\begin{itemize}
    \item \textbf{System message}: You are Meta-Expert, an extremely clever expert with the unique ability to collaborate with multiple experts (such as Expert Problem Solver, Expert Mathematician, Expert Essayist, etc.) to tackle any task and solve any complex problems. Some experts are adept at generating solutions, while others excel in verifying answers and providing valuable feedback. …
    \item Template for the expert prompting: \texttt{f"\{expert identity\}. Now given the above identity background, please answer the following question: Question: \{task description\} \{input\}”}
\end{itemize}

\paragraph{ToT} Imagine three different experts are answering this question. All experts will write down 1 step of their thinking, then share it with the group. Then all experts will go on to the next step, etc. If any expert realises they're wrong at any point then they leave. The question is...

\twocolumn

\section{Dialogues}

\begin{table}[h]
\centering
\resizebox{0.9\columnwidth}{!}{%
\begin{tabular}{L}
\toprule
\textbf{Sonnet Writing} \\ \midrule
\textbf{Ground Truth}: "" \\
\textbf{Question:} ['attacks', 'incredible', 'reduction'] \\
\midrule
\multicolumn{1}{c}{------ DIALOGUE START ------} \\
\midrule
\textbf{Performer}: When twilight shadows cloak the day's retreat, (A) \\
The world, in silence, holds its breath in awe; (B)\\
Yet in the dark, our fears often repeat, (A)\\
And peace is shattered by the claws of war. (B)\\
 \\
Incredible how quickly life can change, (C)\\
From tranquil scenes to storms that rage and roar; (D)\\
Yet in our hearts, a light remains, unstrange, (C)\\
That guides us through each trial we deplore. (D)\\
 \\
Attacks may come in forms both sharp and sly, (E)\\
A reduction of the joy we knew before; (F)\\
But courage calls, its voice cannot deny, (E)\\
That we shall rise, stronger than we swore. (F)\\
 \\
So hold this truth, and let it be your guide, (G)
In darkest times, hope is your surest tide. (G) \\
 \\
\textcolor{blue}{Expert verification: the solution is not correct.} \\
\midrule
\textbf{Teacher}: Invalid Solution. The sonnet provided does not adhere to the specified rhyme scheme ABAB CDCD EFEF GG. The word "war" in line 4 should rhyme with "awe" in line 2, but it does not. Instead, "war" rhymes with words in other quatrains ("roar", "deplore", "before"), which disrupts the intended ABAB CDCD EFEF GG pattern. \\
 \\
Please revise line 4 to ensure it rhymes correctly with line 2, and check that all other lines strictly follow the designated rhyme scheme. \\ \midrule
\multicolumn{1}{c}{---------- 1 -------------} \\ \midrule
\textbf{Performer}: When twilight shadows cloak the day's retreat, (A) \\
The world, in silence, holds its breath in awe; (B) \\
Yet in the dark, our fears often repeat, (A) \\
And peace is shattered by the jaws of law. (B) \\
 \\
Incredible how quickly life can change, (C) \\
From tranquil scenes to storms that rage and roar; (D)\\
Yet in our hearts, a light remains, unstrange, (C)\\
That guides us through each trial we deplore. (D)\\
 \\
Attacks may come in forms both sharp and sly, (E) \\
A reduction of the joy we knew before; (F) \\
But courage calls, its voice cannot deny, (E)\\
That we shall rise, stronger than we swore. (F)\\
 \\
So hold this truth, and let it be your guide, (G) \\
In darkest times, hope is your surest tide. (G) \\ \midrule
\textbf{Teacher}: Valid Solution. \\

\bottomrule
\end{tabular}%
}
\caption{Example of dialogue for Sonnet Writing. The Expert verification prompt is indicated in blue.}
\label{tab:dialogue_sonnet}
\end{table}

\begin{table}[h]
\centering
\resizebox{\columnwidth}{!}{%
\begin{tabular}{L}
\toprule
\textbf{Penguins In A Table} \\ \midrule
\textbf{Ground Truth}: 'Louis' \\
\textbf{Question:} Here is a table where the first line is a header and each subsequent line is a penguin: \\

name, age, height (cm), weight (kg) \\
Louis, 7, 50, 11 \\
Bernard, 5, 80, 13 \\
Vincent, 9, 60, 11 \\
Gwen, 8, 70, 15 \\

For example: the age of Louis is 7, the weight of Gwen is 15 kg, the height of Bernard is 80 cm. \\
Which is the second youngest penguin? \\
\midrule
\multicolumn{1}{c}{------ DIALOGUE START ------} \\
\midrule
\textbf{Performer}: The answer is Vincent. \\
\midrule
\textbf{Teacher}: Invalid Solution. The Performer's answer does not match the correct answer based on the data provided in the table. The Performer should review the ages of the penguins and determine the correct order to accurately identify the second youngest penguin. Please ensure that the ages are compared correctly to identify the correct penguin based on their age. \\ \midrule
\multicolumn{1}{c}{---------- 1 -------------} \\ \midrule
\textbf{Performer}: Could you clarify how the ages should be ordered to determine the ranking of youngest to oldest? \\ \midrule
\textbf{Teacher}: To determine the ranking of the penguins from youngest to oldest, you should list the penguins in order of their ages from the smallest (youngest) to the largest (oldest). This involves comparing the numerical age values provided in the table for each penguin and arranging them accordingly. Once arranged, you can identify the position of each penguin in this sequence (e.g., youngest, second youngest, third youngest, oldest).
\\ \midrule
\multicolumn{1}{c}{---------- 2 -------------} \\ \midrule
\textbf{Performer}: The answer is Gwen. \\ \midrule
\textbf{Teacher}: Invalid Solution. \\

Please review the data provided in the table and ensure that the ages are compared accurately to determine the correct ranking from youngest to oldest. The Performer's answer violates rules 5 and 7.
 \\ \midrule

\multicolumn{1}{c}{---------- 3 -------------} \\ \midrule
\textbf{Performer}: The answer is Louis. \\ \midrule
\textbf{Teacher}: Valid Solution. \\
\bottomrule
\end{tabular}%
}
\caption{Example of dialogue for Penguins In A Table.}
\label{tab:dialogue_penguin}
\end{table}

\onecolumn

\subsection{StrategyQA Dialogue}
\begin{lstlisting}[language=json,firstnumber=1]
    {
        "role": "user",
        "content": "Welcome! You are a strategist tasked with devising a plan to achieve specific targets given a particular scenario.\n\nFirst, review the current scenario:\nHydrogen's atomic number squared exceeds number of Spice Girls?\n\nConsider the various strategies available and evaluate which one maximizes the likelihood of achieving the target described. Be mindful of the rules outlined and ensure compliance:\n\n \{rules\}
        Let us begin. Please provide your first answer to the question. Write the solution in this following format \"The answer is ... \n"
      },
      {
        "role": "system",
        "content": "The answer is no, the square of hydrogen's atomic number does not exceed the number of Spice Girls."
      },
      {
        "role": "user",
        "content": "Valid Solution.\n\nThe proposed strategy correctly identifies and answers the main question: \"Does the square of hydrogen's atomic number exceed the number of Spice Girls?\" The strategy involves the following reasoning steps:\n\n1. **Identify the atomic number of hydrogen**, which is 1.\n2. **Square the atomic number of hydrogen** (1^2 = 1).\n3. **Compare this result to the number of Spice Girls**, which is 5.\n\nThe answer is derived by recognizing that the square of hydrogen's atomic number (1) is less than the number of Spice Girls (5). This strategy adheres to the rules by:\n- Accurately identifying the core sub-question.\n- Explicitly stating the underlying reasoning steps.\n- Leveraging background knowledge about hydrogen's atomic number and the number of Spice Girls.\n- Focusing on relevant evidence and constructing a clear reasoning chain.\n- Justifying the answer based on the facts provided.\n\nThe response also implicitly acknowledges that there is no ambiguity in the question and that all provided information is relevant and within scope. The confidence in the answer is high due to the straightforward nature of the mathematical calculation and the well-known facts used."
      }
\end{lstlisting}

\subsection{GSM8k Dialogue}
\begin{lstlisting}[language=json,firstnumber=1]
{
        "role": "user",
        "content": "Welcome! You are a problem solver tasked with finding solutions to a specific scenario described below.Scenario:Janet\u2019s ducks lay 16 eggs per day. She eats three for breakfast every morning and bakes muffins for her friends every day with four. She sells the remainder at the farmers' market daily for $2 per fresh duck egg. How much in dollars does she make every day at the farmers' market?Analyze the problem and develop a step-by-step solution that achieves the stated objective. Be sure to adhere to the following rules:\n {rules} 
        Let us begin. Please outline your first solution or ask a clarifying question. Write you answer as:\n'The answer is ...'\n"
      },
      {
        "role": "system",
        "content": "$18"
      },
      {
        "role": "user",
        "content": "Valid Solution. The problem solver's proposed solution correctly calculates the amount Janet makes daily at the farmer's market by selling duck eggs. The solution follows the correct sequence of arithmetic operations, accurately extracts relevant information, performs calculations correctly, and handles units appropriately. The final answer is presented clearly and aligns with the problem's requirements."
      }
\end{lstlisting}

\end{document}